# Psychotic Relapse in Schizophrenia: Routine Clustering of Mobile Sensor Data Facilitates Relapse Prediction

Joanne Zhou, Bishal Lamichhane, Dror Ben-Zeev, Andrew Campbell, Akane Sano

## Abstract

### Background

Behavioral representations obtained from mobile sensing data could be helpful for the prediction of an oncoming psychotic relapse in patients suffering from schizophrenia and delivery of timely interventions to mitigate such relapse.

### Objective

In this work, we aim to develop clustering models to obtain behavioral representations from continuous multimodal mobile sensing data towards relapse prediction tasks. The identified clusters could represent different routine behavioral trends related to daily living of patients as well as atypical behavioral trends associated with impending relapse.

### Methods

We used the mobile sensing data obtained in the CrossCheck project for our analysis. Continuous data from six different mobile sensing-based modalities (e.g. ambient light, sound/conversation, acceleration etc.) obtained from a total of 63 patients suffering from schizophrenia, each monitored for up to a year, were used for the clustering models and relapse prediction evaluation. Two clustering models, Gaussian Mixture Model (GMM) and Partition Around Medoids (PAM), were used to obtain behavioral representations from the mobile sensing data. These models have different notions of similarity between behaviors as represented by the mobile sensing data and thus provide differing behavioral characterizations. The features obtained from the clustering models were used to train and evaluate a personalized relapse prediction model using Balanced Random Forest. The personalization was done by identifying optimal features for a given patient based on a personalization subset consisting of other patients who are of similar age.

### Results

The clusters identified using the GMM and PAM models were found to represent different behavioral patterns (such as clusters representing sedentary days, active but with low communications days, etc.). While GMM based models better characterized routine behaviors by discovering dense clusters with low cluster spread, some other identified clusters had a larger cluster spread likely indicating heterogeneous behavioral characterizations. PAM model based clusters on the other hand had lower variability of cluster spread, indicating more homogeneous behavioral characterization in the obtained clusters. Significant changes near the relapse periods were seen in the obtained behavioral representation features from the clustering models. The clustering model based features, together with other features characterizing the mobile sensing data, resulted in an F2 score of 0.23 for the relapse prediction task in a leave-one-patient-out evaluation setting. This obtained F2 score is significantly higher than a random classification baseline with an average F2 score of 0.042.

## Conclusions

Mobile sensing can capture behavioral trends using different sensing modalities. Clustering of the daily mobile sensing data may help discover routine as well as atypical behavioral trends. In this work, we used GMM and PAM-based cluster models to obtain behavioral trends in patients suffering from schizophrenia. The features derived from the cluster models were found to be predictive for detecting an oncoming psychotic relapse. Such relapse prediction models can be helpful to enable timely interventions.

# Introduction

## Background

Schizophrenia is the most common psychotic disorder, affecting up to 20 million people worldwide [1] and accounting for more than 13.4 million years of life lived with a disability [2]. It can be caused by a combination of genetic, environmental, and psychosocial factors. Patients with schizophrenia experience ranges of positive symptoms (hallucinations, delusions, etc.), negative symptoms (anhedonia, social withdrawal, etc.), and cognitive dysfunctions (lack of attention, working memory, executive function, etc.) [3,4]. The disorder is highly disabling and often has consequences such as impairment of education, employment, and social contact [4]. Adults with schizophrenia also have an increased risk of premature mortality than the general population [5]. Proper treatment and management of schizophrenia are therefore important to limit the negative life impact of the disorder.

Schizophrenia is usually treated with a combination of antipsychotic medications and psychosocial treatments. However, patients under treatment can still experience psychotic/symptomatic relapse, an acute exacerbation of schizophrenia symptoms [6]. A prior study found that the cumulative first and second relapse rate was 81.9% and 78% respectively within 5 years of recovery from the first episode of schizophrenia and schizoaffective disorder [7]. The risk of relapse is found to be significantly higher after treatment reduction or discontinuation [6]. Relapse poses severe health risks for the individual and can jeopardize their treatment progression and daily functioning. Each relapse episode is associated with a risk of self-harm and harm to others [8].

To keep track of a patient's health status and recovery, routine clinic visits for continual assessment are recommended. Clinical interview and questionnaire tools are used during the visit for assessment of current health symptoms and timely intervention to prevent relapses [9]. However, relapses may happen between the visits during which a patient's health status is not assessed. In addition, patients may have limited insight during a psychotic relapse and struggle to report it to the treatment team or a significant other. Therefore, improving treatment adherence and preventing relapses have become a focus of schizophrenia management. Towards the effort of relapse prevention, there has been significant interest in mobile sensing-based behavioral monitoring models for automatic relapse risk prediction.

## Prior Work

Smartphone apps and wearable devices have been employed in several previous works to collect passive sensing data and track daily behaviors, which could then be used to model the relationship between behaviors and mental well-being. For example, in the Studentlife study, an Android sensing app collected passive sensing data from 48 college students and the inferred behavioral features from the collected data were found to be correlated with academic performance and self-reported mental health conditions [10]. In a study on depression severity, the mobile sensing-based features such as daily behavioral rhythms, variance of subject's location, and phone usage were found to be related to depressive symptom severity [11]. The use of mobile sensing to collect long-term monitoring data has also been demonstrated to be feasible and acceptable for patients with schizophrenia disorders [12–15]. Surveys have found that people with schizophrenia commonly access digital devices for communication and support related to the disorder, which again shows the applicability of using mobile sensing as a platform to monitor schizophrenia symptoms [16].

Mobile sensing data has been used to model behaviors and predict psychotic relapses of patients suffering from schizophrenia. If an oncoming relapse could be detected with high accuracy, then timely medical interventions could be provided to mitigate the associated risks. Researchers have found anomalies in daily behavior assessed from mobile sensing before relapses and developed relapse prediction models with promising accuracy [17–19]. In a pilot study, the Beiwe app collected mobile sensing data from 15 patients suffering from schizophrenia for 3 months during which 5 patients experienced relapses [17]. The researchers found that the rate of anomalies in mobility and social behavior increased significantly closer to relapses. In the CrossCheck project, a mobile sensing app was developed to collect self-reporting EMA (Ecological Momentary Assessment) and continuous passive sensing data from 75 outpatients with schizophrenia [20]. Based on this dataset, the authors in [18] compared different machine learning models for relapse prediction, with several feature extraction windows, and identified the best classifier and prediction settings for detecting an oncoming relapse. The best performance was obtained using an SVM (with RBF kernel) model and a feature extraction window of 30 days, leading to an F1 score of 0.27 on the relapse prediction task. Similarly, the authors in [21] used an anomaly detection framework based on an encoder-decoder reconstruction loss to predict psychotic relapse in schizophrenia.

Concerning current mental health status, the extent to which an individual adheres to work, sleep, social, or mobility routine, i.e. a regular behavioral pattern, largely impacts their mental well-being and symptom severity of mental disorders [11,22,23]. Behavioral stability features that measure the adherence to routines have been proposed as relapse predictors in some of the previous studies. Features computed in our previous work measured behavioral stability by calculating the temporal evolution of daily templates of features derived from the mobile sensing data (daily templates are time-series obtained with representative feature values at regular time-intervals in a given day, e.g. time-series of hourly feature values) [19]. The authors in [24] also showed the effectiveness of using behavioral rhythm-based features to predict different symptom severity. Stability features such as deviation of daily templates were found to be significant predictors of schizophrenia symptoms such as being depressed. The authors in [25] also proposed a stability metric for behaviors with a fine temporal resolution by calculating the distance between two cumulative sum functions describing behaviors in a certain minute of the day. The computed Stability Index had similar predictive power as the state-of-the-art behavioral features (mean and standard deviation of each behavior) in [26], while being complementary. In all of these previous works utilizing behavioral stability to model relapse prediction, the stability measured was limited to the behaviors observed within a short feature extraction window (e.g. few weeks only). An individual's routine behaviors were not fully represented due to the short time window considerations. A summary of behavioral patterns could rather be obtained when larger time windows are considered.

In this work, instead of measuring behavioral patterns using the variance of day-to-day behaviors, we identify the overall cluster of behaviors for an individual using multimodal mobile phone data and unsupervised machine learning, and derive features based on the distance of behaviors observed in a day compared to the individual's most representative routines. The identified behavioral clusters for an individual could for example be representing their weekday routine, a weekend routine, and a low-phone-usage routine (no sensor reading), etc. The clusters identified provide a representation of the long-term behavioral trends across the subjects which are not directly captured by short-term behavioral rhythm features as used in previous works. Further, clusters obtained from the mobile sensing data represent quantized behaviors, and features derived from these clusters are robust to the insignificant variations in behavior compared to the short-term behavioral rhythm change features. Typical behavioral routines for an individual can be found via the clustering analysis of their daily behaviors. Previously, clustering has been applied for identifying mobility patterns using GPS sensing data and evaluating anomalies accordingly [21,26]. However, to the best of our knowledge, clustering analysis hasn't been done for characterizing the overall behavioral patterns of patients with schizophrenia, using multi-modal mobile sensing data, towards relapse prediction tasks.

## Goal of This Study

In this work, we aim to (1) develop a method to characterize patients' daily behaviors using multimodal smartphone sensor data, (2) understand the relationship between behavioral patterns and psychotic relapse events in schizophrenia, and (3) evaluate the predictive power of identified behavioral pattern-based features for relapse prediction. We propose multivariate time-series clustering of daily templates obtained from mobile sensing data to obtain behavioral patterns. The features derived from clustering are then used in the relapse prediction task. The paper is organized as follows. In the Methods section, we describe the method used to cluster multi-dimensional daily templates from mobile sensing data, model selection approach for clustering, as well as feature extraction and relapse prediction modeling. In the Results section, we present the results obtained from the clustering models, association of the obtained clustering-based behavioral features with relapses, and evaluation of the developed relapse prediction model. The obtained results are discussed, and future directions are outlined in the Discussions section.

# Methods

## Data Preparation

The data used in this study was obtained from the CrossCheck project (Clinical Trial Registration Number: **NCT01952041** [27]), which was conducted at the Zucker Hillside Hospital in New York City [20,24,26,28,29]. The study was approved by the ethical review committee at Dartmouth College and the institutional review board at North Shore-Long Island Jewish Health System [20]. Informed consents were obtained from the participants. The inclusion criteria for the participant has been described in [20]. The CrossCheck app collected mobile sensing data from 75 outpatients with schizophrenia, with a data collection period of over 12 months per patient. Sixty-three patients completed the data collection (27 male and 36 female, average age of 37.2 +/- 13.7 years, minimum age 18 years, and maximum age 65 years), and a total of 27 relapse events occurred in 20 patients during the monitoring period. Some patients had multiple incidences of relapses but as the monitoring period was long, each of the incidences was treated as a unique event if separated by a month. A relapse incident was defined to have occurred under one or more of the following seven different criteria: psychiatric hospitalization, increased frequency or intensity of services, increased medications or dosages or over 25% changes in BPRS scores, suicidal ideation, homicidal ideation, self-injury, and violent behavior resulting in harm to self or others [18]. Six mobile sensing modalities including physical activities, sociability, and ambient environmental readings were obtained using the app. Different features were extracted from these mobile sensing modalities as presented in [24]. From among these features, a total of 21 passive sensing features were selected for our proposed clustering-based behavioral characterization: acceleration, distance traveled, sleep duration, ambient sound, ambient light, conversation duration, phone unlock duration, and different types of call log, sms log, and app usage. All the features were transformed to an hourly resolution, by averaging the observations within one hour. For features that were obtained with lower resolution (e.g. every few hours), for example, distance traveled from morning to noon, the feature values were split to each hour spanned by the time represented by these feature values. With hourly resolution for each of the 21 features considered, and these hourly feature values considered as separate feature space, the resulting dataset had a dimension of 504 (21 x 24). A total of 18436 days of observation are present from the data collected for all the patients. Per-patient feature normalization (min-max normalization between 0 to 1) was done to adjust for differences between patients. From the normalized dataset, principal components analysis (PCA) on the full dataset (with data from all the patients) was done for dimensionality reduction. The first 200 principal components were retained which explained 96.9% of the total variance.

## Clustering Models

We evaluated two different clustering methods: Gaussian Mixture Model (GMM) and Partitioning Around Medoids (PAM), to cluster the features from the mobile sensing data and obtain behavioral representations. The two clustering models differ in how the similarity between different points are assessed, representing different ways in which behaviors across days can be compared to each other, and therefore produce different cluster outputs.

### Gaussian Mixture Model

#### Model Introduction

The GMM is a probabilistic model that assumes data is generated from a finite set of Gaussian distributions. A Gaussian mixture probability density is the weighted sum of *k* component Gaussian densities [30]. The GMM model can address correlation between attributes by selecting the optimal covariance matrix for each cluster and has been employed in previous behavioral clustering problems [31]. Moreover, it can derive the probability of each sample in its assigned gaussian distribution. In this study, we used the GMM implementation from the scikit-learn package in Python to obtain a clustering model for the mobile sensing data [32]. The parameters of the GMM model were obtained using the expectation-maximization (EM) algorithm [33]. We selected the number of clusters and the covariance matrix type based on Akaike information criterion (AIC) and Bayesian information criterion (BIC) scores of all the candidate models (See more details in the supplementary document).

#### Model Output

Three output variables for each of the data points (observations), offering GMM model-based clustering features for the data points, are generated based on the developed GMM model: cluster label, assigned cluster likelihood score, and weighted average likelihood score.

Cluster label is represented by integers from 1 to *k (k: number of clusters selected in the GMM model)*. Cluster likelihood scores derived from the model measure how "irregular" each day (represented by a data point) is by calculating its deviation from the Gaussian mixtures. If we consider the center of each of the Gaussian as a typical routine, then the farther out a point is in this Gaussian space, then higher the chances that the point represents an anomalous day/behavior are.

The likelihood of a data point in a multivariate Gaussian distribution can be computed by calculating the probability of observing a point farther than this given point. In other words, the cumulative distribution function is evaluated at the given data point, which can be obtained using Mahalanobis distance metric. Note that the squared Mahalanobis distance from a point to the center of a Gaussian distribution has been proven to follow a chi-squared distribution with $p$ degrees of freedom, where $p$ is the number of variables [34]. Therefore, the likelihood of a point in the Gaussian distribution is equivalent to the cumulative probability of observing a value larger than the given Mahalanobis distance in a chi-squared distribution with $p$ degrees of freedom.

The assigned cluster likelihood score of the data point was obtained as the probability of each point to its assigned cluster. The weighted average likelihood score was computed as the weighted (with the cluster's corresponding weights) sum of the probability of a given point belonging to each of the Gaussian classes. Intuitively, the assigned cluster likelihood score measures how close a day is to its closest routine. The weighted average likelihood score measures how close a day is to all routines. Since the weighted average likelihood score accounts for cluster weights, a point that is closer to a more populous cluster will be considered less anomalous. A 2-D illustration of the likelihood scores is provided in Supplementary Figure 1.

## Partition Around Medoids with Dynamic Time Warping

### Model Introduction

GMM models measure similarity between observations (data points) with point-wise alignment of different features in the observation. However, the dissimilarity between two observations could be overestimated due to an outlier (e.g. because of faulty sensor measurements) or when there is a small time-shift and/or speed difference between observations. For example, two daily templates with a similar pattern but a shift by one hour would be expected to represent similar behavioral representations but these templates would likely be considered dissimilar from a GMM model. To allow flexible similarity assessments, we used Dynamic Time Warping (DTW) to find the optimal alignment of indices of the two time-series that minimizes the distance between the time-series [35]. The DTW distance can be paired with a distance-based clustering method, such as a partition around medoids (PAM) clustering model [36]. The PAM model searches for *k* representative objects (medoids) from the data and creates clusters so that the total dissimilarity of points within clusters is minimized. We compared the number of clusters k based on the sum of the squared DTW distance of every data point to its cluster medoid and the elbow method (See more details in Supplementary document).

## Model Output

From the fitted PAM model, similar to the procedure after GMM model fit, we generated three output features characterizing each data point: cluster label, assigned cluster distance score, and weighted average distance score. As in the GMM model-based likelihood score computation, the cluster distance scores evaluate how dissimilar each object is from a representative data point, or from all representative data points. The assigned cluster distance score is the DTW distance of each data point (representing a daily template) to its cluster medoid. A lower value means that a day conforms better to its closest routine. Weighted average distance score is obtained by summing the DTW distance to all medoids scaled by the corresponding cluster sizes. A lower value means that a day conforms better to all possible routines. DTW distance from the previous day's daily template was also calculated as a potential relapse predictor.

## Analyzing Cluster Results

After obtaining output variables from the cluster models, we evaluated whether there were significant changes in any of these cluster output variables closer to relapse events. To quantify this change, we first defined different key periods to focus before a relapse. Similar to a previous work, we defined NRx as x days near relapse (before the relapse event) and pre-NRx as all days before relapses that are not in NRx (healthy period) [21]. We evaluated cluster outputs for NR7, NR14, NR20, and NR30 periods to test different window sizes. Cliff's delta was computed to estimate the size of the change in the likelihood scores (GMM model output) and distance scores (PAM model output) between the NRx and pre-NRx periods for each patient separately [37]. Cliff's delta was chosen because of the non-normality and variance heterogeneity of our data for which the Cliff's delta is a suitable metric. It is calculated as

$$\hat{\delta} = \frac{\#(x_1 > x_2) - \#(x_1 < x_2)}{n_1 n_2}$$

where $\#(x_1 > x_2)$ counts the number of values in group 1 (NRx period) that is larger than a value in group 2 (pre-NRx period) for all value pairs, and $n_1$ and $n_2$ are the sample sizes. This effect size ranges from -1 to 1, where 1 indicates that all values in the NRx period are larger than all values in the pre-NRx period, and -1 indicates vice-versa. As proposed by Romano et al., the effect can be considered to be non-negligible if the absolute value is larger than 0.147 [38]. Cliff's delta is suitable to compare continuous variable output such as likelihood scores and distance scores.

# Relapse Prediction

## Relapse prediction approach

We framed relapse prediction as a binary classification problem similar to the earlier works [19,39]. Based on the mobile sensing features derived from a feature extraction window (current and immediately past observations from a patient), we predicted if the patient is likely to experience a relapse in an oncoming period (prediction window). Similar to the previous works [19,39], we used a 4-week period as the feature extraction window and a 1-week period as the prediction window (Figure 1). Thus, the mobile sensing observations from the past 4-week period are used in the relapse prediction model to predict if there is going to be a relapse in the next week.

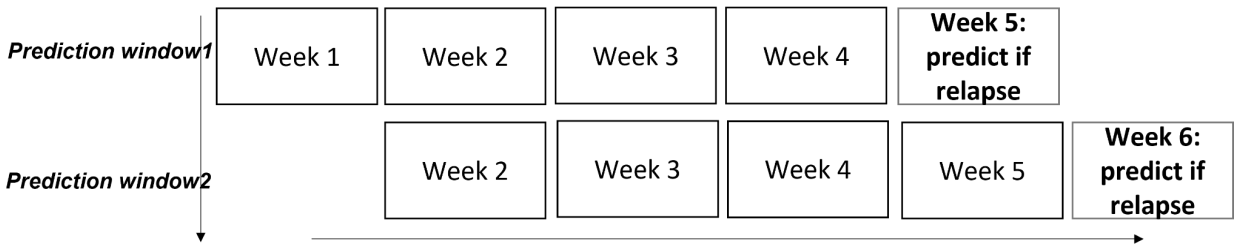

Figure 1: Sequential relapse prediction approach used in this work. Features are extracted from a period of 4 weeks in order to predict if there is likely to be a relapse in the coming week.

## Features

Mobile sensing data are represented with features to characterize behavioral patterns in the relapse prediction model. For our work, we evaluated the contribution of the clustering features derived from the GMM and PAM models for the psychotic relapse prediction task. We briefly describe the baseline features (based on the earlier work [19]) and clustering-based features that are added for the relapse prediction model.

1. Baseline features:
These consist of all the features as used in [19] along with distance-based and duration-based mobility features as well as screen usage-based features. The crosscheck dataset contains information about when the screen of the subject's smartphone is active. A single screen-usage modality was derived that represents the time spent using the phone (phone screen was active). From this modality, the mean and standard deviation of daily averages in a given feature extraction window was computed as features for the relapse prediction model. Similarly, for mobility-based features, we computed four different mobility modalities: distance traveled from home (home information obtained based on the clustering of the GPS locations), total movement, average time stayed in a location, and total time spent at home. Then for each mobility-based modality, we computed the mean and standard deviation of the daily averages as features characterizing a feature extraction window.

2. Clustering-based features:
We extended the baseline feature set with our proposed clustering-based features for the relapse prediction task. These features are listed in Table 1.

Table 1. Features used in relapse prediction models. Baseline features are derived from a previous work [19]. We evaluated if the clustering-based features could improve relapse prediction by complementing the daily behavioral rhythm change based features represented in the baseline features.

| Feature Set | Modalities | Features |
| --- | --- | --- |
| Baseline features | Accelerometer magnitude<br>Ambient light<br>Distance traveled<br>Call duration<br>Sound level<br>Conversation duration | Mean daily template features (*mean, std, max, range, skewness, kurtosis*) Standard deviation template features (*mean*), Absolute difference between mean and maximum template (*max*), distance between normalized mean templates, weighted distance between normalized mean templates, distance between normalized maximum template and mean template, daily averages (*mean, std*) |
| | 10-item EMAs | *Mean* and *standard deviation* of EMA items in feature extraction window |
| | Screen usage<br>Distance-based mobility features<br> - Distance from home<br> - Total movement<br>Duration-based mobility features<br> - Time in a location<br> - Time spent at home | *Mean* and *standard deviation* of daily averages in feature extraction window |
| Clustering features | GMM features | *Mean* and *standard deviation* of (GMM label, GMM |

|  |  | likelihood scores), number of cluster transitions, number of cluster states |
| --- | --- | --- |
|  | PAM features | *Mean* and *standard deviation* of (PAM label, PAM distance scores, DTW difference from the previous day), number of cluster transitions, number of cluster states |
| Demographic features |  | Age, Education years |

### Classifier

For our relapse prediction pipeline, we used a Balanced Random Forest (BRF) classifier with a low overall model complexity (using 11 decision trees). BRF as a classifier is suitable for learning from an imbalanced dataset, as is the case in our relapse prediction task, and provides meaningful prediction probabilities in different decision fusion schemes (e.g. in situations where only a limited number of sensor modalities are available for a patient). The number of decision trees to be used was heuristically chosen so as to limit model size (lower number of trees) while still having a number of trees to maintain diversity for the generalizability of the ensemble model. We used the BRF implemented in the imbalanced-learn library in Python [40] allowing the default unrestricted depth of trees and *sqrt(number of features)* considered for best split in the trees. Similar to the approach used in [19], features are quantized into discrete bins before being provided as input to the classifier. The number of bins is set as a hyperparameter and for the set number of equal-width bins, the count of feature values in each of the bins are retained as the processed feature values. The approach of feature quantization was found to be helpful in relapse prediction, probably by blunting small insignificant changes while retaining larger feature variations representing significant behavioral deviations. We used leave-one-patient-out cross-validation for the evaluation of the model. The number of bins to be used is a hyperparameter for the classification model and was set with cross-validation within the training set (nested cross-validation). The number of bins for feature quantization considered in hyperparameter tuning were [2,3,4,5,10,15] and the tuning procedure is further described in the supplementary file.

### Relapse Labels

For our relapse prediction pipeline, as the relapse dates are not a hard label and earlier indications of an oncoming relapse are also valuable, we regard the entire month preceding the date of indicated relapse as a relapse period for classification. Thus, any prediction of relapse within a 4-week period before the relapse is considered as an useful output from the prediction model, as has also been used in previous work on relapse prediction [21]. A relapse prediction generated upto a month before the relapse would be observable and potentially actionable for interventions as behavioral changes associated with relapse could manifest up to a month preceding a relapse [18].

## Personalization

Human behavior and behavioral change manifestations of relapse could be person-dependent. The authors in [19] proposed a method for personalizing a relapse prediction model based on feature selection adapted to a particular test patient. The adaptation happens using a personalization subset. This is shown in Figure 2. For a test subject, within the leave-one-patient-out cross-validation approach, the data from subjects closest in age to the given test subject compose the personalization subset. We included age-based personalization as a first step towards personalized relapse prediction since behavioral tendencies could be dependent on age, among other factors. Age has been reported to be a significant factor in univariate regression modeling of relapse behaviors in patients suffering from schizophrenia [41] and age dependence of psychosocial functions, substance use behaviors, psychotic symptoms, hospitalization risks, etc. have been reported in the context of psychotic relapse in patients suffering from schizophrenia [42]. We evaluated the gains from age-based personalization compared to a non-personalized model to establish empirically if age-based personalization could be helpful in behavioral modeling and relapse prediction. As the relapse incidents are rare, all the relapse incidents in the training dataset are included as a part of the personalization subset.  For training a classifier towards the test subject, the optimal features are selected using the personalization subset. We employed this approach for training our relapse prediction model and used the correlation between features and target label as the feature selection criteria. The number of features to be selected is set as a hyperparameter in our classifier and this dictates the threshold on correlation value used for feature selection. For example, if the number of features to be selected is 5, then the threshold on correlation coefficient (absolute value) is selected such that top-5 features with the highest correlation with the labels are retained. The number of features to be used was selected from [ 3, 5, 10, 15] features and the size of the personalization subset was selected from [ 50, 75, 100, 125, 150, 200, 300] in the hyperparameter tuning (further described in the supplementary file).

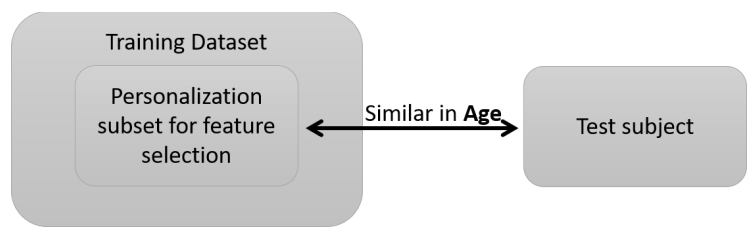

Figure 2: Personalization approach for the relapse prediction model, as proposed in [19]. A personalization subset, consisting of data from subjects who are closest in age to the test subject, is used to identify the best feature sets using which then a (personalized) relapse prediction model can be trained.

### Evaluation Metric

We evaluated relapse prediction performance to assess the contributions from clustering-based features. Any improvement in the relapse prediction performance when clustering-based features are added to the baseline features would establish the value of clustering-based features to represent behavioral trends and detect anomalies relevant for relapse prediction. Similar to [19], we used F2 score for model evaluation to slightly prioritize recall over precision (Detecting a relapse is slightly prioritized over generating a false positive). F2 score is given as:

$$F2 = \frac{5 * precision * recall}{4 * precision + recall}$$

We also report the obtained precision and recall scores together with the F2 scores.

## Results

### Clustering Results

We trained GMM and PAM models to obtain cluster centers and identify different behavioral routine representations. The model selection procedure is explained in the supplementary file, and model comparison metrics for GMM and PAM are plotted in supplementary figure 2 and 3. For the GMM model, after evaluating AIC and BIC scores, model selection was narrowed to 8 to 14 clusters with full covariance matrix. Among the models with equally good AIC and BIC scores, the models with 9 and 13 clusters achieved the best model stability and least overlap between Gaussian classes. The final model selection was 9 clusters because lower number of clusters allows for higher interpretability. The number of clusters for the PAM model was also selected to be 9 based on the distance dissimilarity metric and the elbow method.
See supplementary Figures 4 and 5 for the output from the GMM and PAM models including cluster size, average likelihood scores (for GMM), and distance scores (for PAM) (Figure S4) and kernel density plots that illustrates the distributions of likelihood scores and distance scores (Figure S5).

To evaluate how well the days in each cluster conform to one routine - the one represented by the cluster center - we measured the spread of each cluster using the trace of the covariance matrix of all cluster samples. Results are illustrated in Figure 3. Clusters with a smaller covariance trace have lower within-cluster variability. The GMM cluster model resulted in a more extreme distribution of cluster spread (higher range of covariance trace) because it allows the clusters to overlap (despite our model selection approach to limit overlaps) while the PAM model creates partitions in the data.

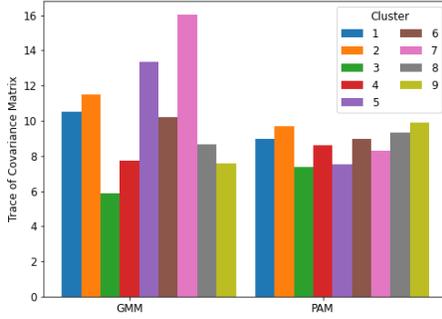

Figure 3: Trace of the sample covariance matrix for each cluster obtained with GMM and PAM clustering approach. A lower covariance matrix trace indicates more homogeneous clusters, i.e. clusters with lower within-cluster variability.

By averaging all daily templates (data points) in every cluster, it is possible to observe the cluster profiles. For example, Figure 4 illustrates the average daily templates of two example signal modalities: acceleration and volume. The GMM model performs better in stratifying daily templates based on the overall level of activity in these signal modalities. The PAM model has higher variance in each cluster because it allows for a more lenient dissimilarity measurement. Although the daily templates in each cluster have different levels of signal activity, they generally follow the same pattern as a normal circadian rhythm, e.g. the volume signal activity peaking during the day and being at minimum during the night.

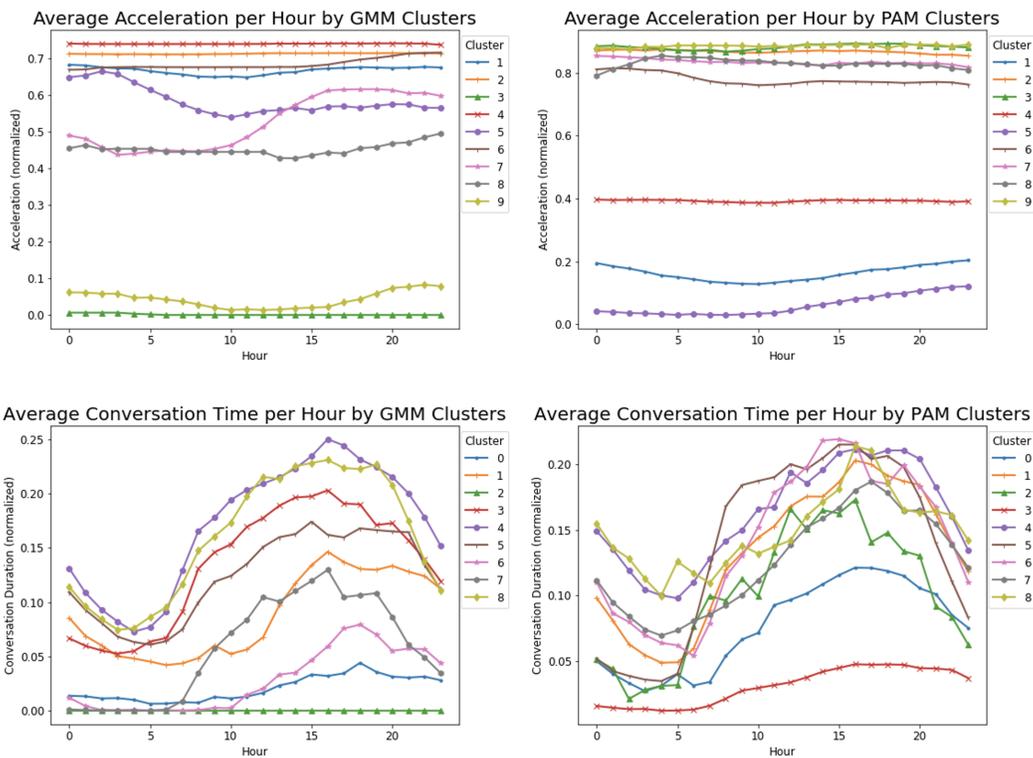

Figure 4: Average daily templates of two signal modalities acceleration (top) and volume (bottom) in the clusters obtained from the GMM and PAM models. Different clusters capture different behavioral patterns.

Table 2 summarizes the average profile for each cluster, ordered from the most common to the least common one.

Table 2. All cluster profiles obtained from the GMM and PAM models in descending cluster size. Different clusters are associated with peculiar behaviors specific to that cluster as it can be observed from the typical profile of signal modality in that cluster.

| Cluster size rank | GMM cluster profile | GMM cluster size | PAM cluster profile | PAM cluster size |
|---|---|---|---|---|
| 1 | No app usage; high conversation and sms; other attributes around average | 5217 | Low acceleration, conversation, volume, sleep duration; Very low variability in sleep and volume templates | 3318 |
| 2 | Highest app usage and phone calls; high acceleration, conversation, sms, distance moved and volume; early wake-up (around 7am) and no sleep during the day | 3993 | High volume and sms; Constantly low sleep template | 3300 |
| 3 | Almost all sensor readings near 0 | 2580 | Conversation and volume sharply increase after 6am; highest volume; low phone usage before 7am; wake-up around 7am and sleep around 9pm | 2728 |
| 4 | Highest acceleration; low phone calls; wake up early (around 7am) and no sleep during the day | 1883 | High app usage, sms, and distance moved around midnight; below average acceleration | 2699 |

| 5 | High acceleration after midnight; high phone calls and sms; high overall volume even at night; late sleep and wake-up | 1484 | Lowest acceleration (close to 0) and app usage; constantly high screen time and sleep duration | 2378 |
| --- | --- | --- | --- | --- |
| 6 | Below average volume and distance; wake-up after 11am and sleep during the day | 1298 | High phone call and sms; screen time sharply increases after 6am; wake-up around 9am and sleep around 11pm; awake during the day | 1686 |
| 7 | Activity level and phone usage are highly active during the day and inactive at night; short sleep duration; high number of phone calls; acceleration increase after 3pm | 1046 | Below average screen time; long sleep time (wake up around noon) | 1405 |
| 8 | No app usage; low conversation, sms and volume; long sleep even during the day | 523 | Low phone call, sms and screen time; high volume at night; constant long sleep (wake up in the afternoon) | 752 |
| 9 | Accelerometer readings close to 0; low app usage, conversation, and volume; phone screen is constantly on; long sleep duration even during the day | 412 | High app usage and distanced moved around noon; templates in this cluster have high dissimilarities | 170 |

## Association with Relapses

Out of the 27 relapse events in total, clustering features were missing before three events due to missing signal modalities. For 11 out of the 24 relapses left, anomalies in clustering features were observed qualitatively in the time series of these features before and after the relapse. Most of these anomalies represent a transition to a cluster with inactive sensor readings, for example, GMM cluster 3 and PAM cluster 1 (Figure 5). We hypothesized that these patients for which we see their assigned cluster labels near relapse period being assigned to the cluster of inactive sensor recordings, most likely had their phone turned off a few days before the relapse. This transition to an inactive cluster is associated with an increase in likelihood scores (GMM model-based feature), and a decrease in distance scores (PAM model-based feature), because these clusters are more compact, and points do not deviate too much from the cluster centers.

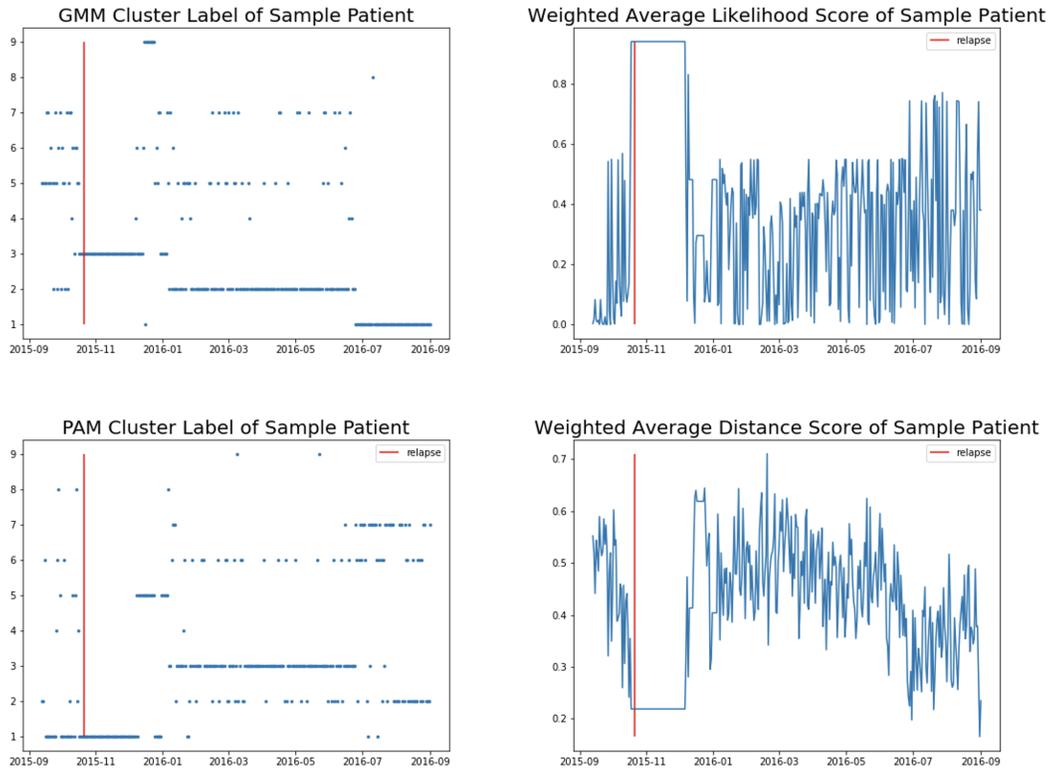

Figure 5: Time series plots of cluster assignment as obtained from the GMM and PAM models (left pane), and weighted average likelihood score and distance score of a sample patient (right pane). Changes in cluster features are seen near to the relapse instance (here shown with the vertical red line).

Our cluster analysis between the NRx and pre-NRx periods showed that on average, likelihood scores increase, and distance scores decrease closer to relapses (Figure 6). This trend is robust with respect to different window sizes and the largest change is observed with NR20 window size. Asterisks indicate that the absolute Cliff's Delta value between the two periods is above 0.147 (i.e. effect is non-negligible, Ref: Section Methods – Analyzing Cluster Results). Note that the plots are made with all patients' data collectively. Individual patient's plots would show a larger difference between the near relapse window and healthy period. Average cliff's delta values across all relapse events are presented in Table S1 in Multimedia Appendix 1.

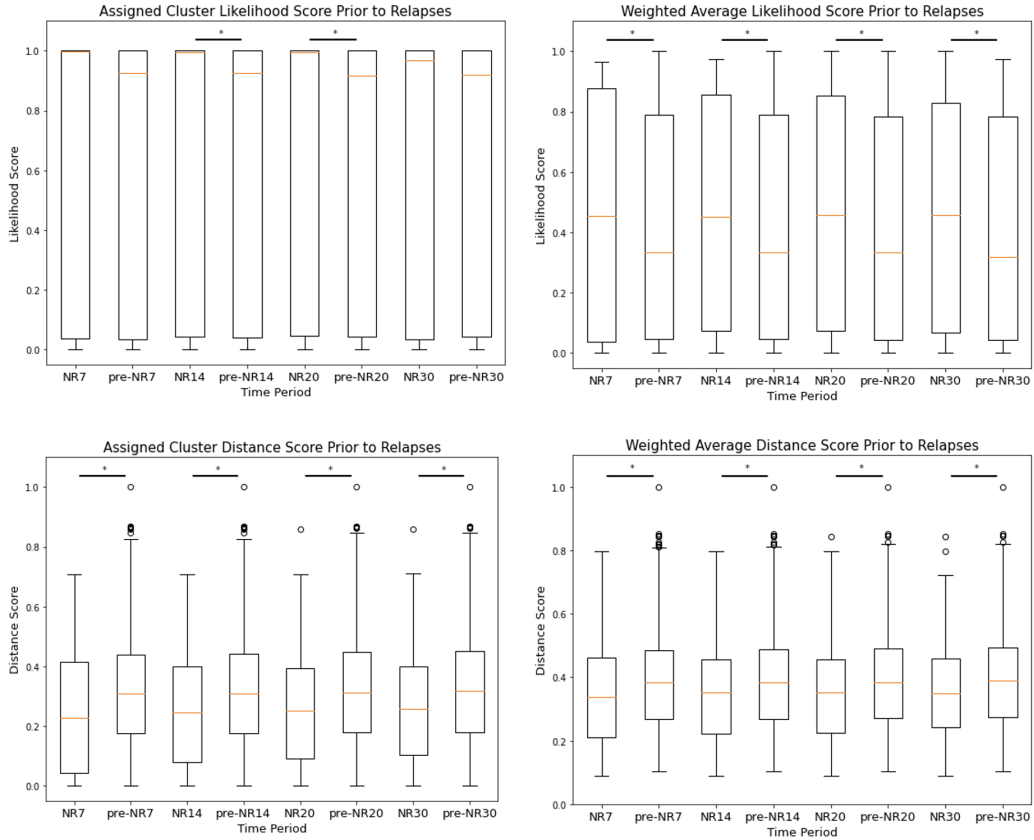

Figure 6: Boxplot of the clustering features (likelihood scores from GMM model on top and distance scores from PAM model on bottom) in different NRx (x days near relapse) and pre-NRx (all days before relapses not in NRx) periods. Asterisks indicate that Cliff's Delta between two groups is above 0.147.

## Relapse prediction

We evaluated the relapse prediction pipeline discussed in the Methods - Relapse Prediction section, with and without the clustering-based features. The highest F2 score of 0.23 is obtained when the baseline features are complemented with the clustering-based features, significantly higher than the random classification baseline of 0.042 F2 score and the F2 score of 0.18 obtained using the baseline features only.

Table 3. Relapse prediction performance with different feature sets. The baseline features introduced in the previous work by [19] are complemented with clustering-based features for evaluation. The performance of both the GMM-based and PAM-based feature sets are also separately evaluated.

| Method | F2 score (precision/recall) |
| --- | --- |
| All features | **0.23 (0.063/0.662)** |
| Baseline features [19] | 0.18 (0.055/0.400) |
| Clustering features | 0.14 (0.035/0.487) |
| GMM features | 0.16 (0.042/0.487) |
| PAM features | 0.19 (0.042/0.525) |
| GMM + Baseline features | 0.19 (0.052/0.525) |
| PAM + Baseline features | 0.16 (0.045/0.438) |
| Random Classification Baseline | 0.042 +/- 0.020 |

Significant features

With the best relapse prediction obtained using all features, we identified the most important features within this feature set based on how often a feature was selected in the leave-one-patient-out cross-validation. The selection count for a feature was incremented by 1 if it was selected for use in a particular cross-validation loop for a test patient. It is to be noted that the number of features selected in each cross-validation loop is different since the number of features is a hyperparameter selected with nested cross-validation. We then normalized the total selection count of each feature at the end of the cross-validation by the number of cross-validation loops. The results obtained are given in Table 4.

Table 4. The top-10 significant features in the relapse prediction pipeline based on the entire feature set (baseline and clustering-based features). The frequency of selection of a particular feature across the cross-validation loop is used to assess the most significant features for relapse prediction. It is to be noted that different numbers of features are selected in each cross-validation loop since the number of features to be used is a hyperparameter tuned with a nested cross-validation loop.

| Features | Frequency (normalized) |
|---|---|
| Baseline feature - distance template skewness | 0.19 |
| Clustering feature - mean PAM label | 0.17 |
| Clustering feature - mean PAM weighted distance | 0.14 |
| Baseline feature - conv template skewness | 0.14 |
| Clustering feature - Number of transitions | 0.12 |
| Clustering feature - Standard deviation GMM label | 0.10 |
| Clustering feature: Standard deviation PAM label | 0.10 |
| Clustering feature: Mean GMM label pval | 0.10 |
| Baseline feature - Conv. template kurtosis | 0.08 |
| Baseline feature - Volume template range | 0.08 |

## Discussions

### Principal Results

In this work, we used clustering models to obtain behavioral representation from mobile sensing data which could be useful for relapse prediction. The two clustering models explored in this study, GMM and PAM, grouped observations using different notions of distance/similarity between data points and therefore captured different behavioral representations (Table 2, Figure 4). These representations can be useful in downstream applications such as relapse prediction.

The GMM model defines distance based on one-to-one matching between the hourly observation of mobile sensing data in the daily template. The clusters identified from the GMM model have a widely varied distribution of cluster spread (Figure 3). With some compact clusters (represented by low cluster covariance) being identified within the GMM model, the rest of the data points that do not belong to any of these compact clusters are considered as a large-spread cluster with no typical cluster profile. These large-spread clusters contain the compact clusters also (cluster overlaps); a point belonging to compact clusters also shows high likelihood of belonging to the large-spread cluster. As we wanted the clusters to capture distinct behavioral trends, we evaluated Bhattacharyya distances to identify the best clustering model with least overlap between identified clusters. The PAM model with DTW distance allows a more lenient match of daily templates of behaviors as represented by the mobile sensing-based features. Such a lenient matching fits the context of this study since DTW is able to discount spikes, speed differences or time shifts when evaluating dissimilarity between two daily templates of behaviors. However, the clusters obtained from the PAM model contain more dissimilarity. It is then more difficult to summarize the cluster profiles for qualitative model interpretation.

Overall, GMM based modeling is able to identify highly dense/populous clusters with very specific behavior associated with these clusters and some dispersed clusters that do not have a typical cluster profile. For example, cluster 3 and cluster 9 identified from the GMM model (Table

2) represent two types of typical routines. Cluster 3 from the GMM model has almost all sensor readings close to zero other than sleep, likely representing an inactive/sedentary day, and cluster 9 has the days with the phone screen always turned on, likely representing a day with high mobile phone usage. The PAM model also has a cluster with mostly inactive days and constantly long screen time (cluster 5). However, this cluster has higher cluster variance. When the average cluster profile of this cluster is observed (Figure 4), some days that do not strictly follow these patterns of inactive day and long screen time are also assigned to the cluster. In terms of behavioral features, this implies that clusters obtained from a PAM model are likely to cluster together behaviors that do not always show homogeneity based on qualitative observations. This is because of the flexibility in the PAM model in allowing unparalleled alignment between behavior time-series. Nonetheless, it might be beneficial to consider PAM based modeling for previously mentioned features: ability to discount spikes and speed differences or time shifts when evaluating dissimilarity between two daily templates of behaviors. Similarity (or dissimilarity) between behaviors might not always be fully represented by hourly alignment and comparison of mobile sensing data across days.

The behavior of a particular day, represented by the mobile sensing data template for that day, was characterized in a clustering model with different clustering-based features such as Gaussian likelihood and DTW distance to the cluster centers. For days with assigned cluster likelihood scores close to 1 and assigned cluster distance scores close to 0, they tend to belong to a dense cluster with a small spread. For example, cluster 3 from the GMM model and cluster 5 from the PAM model respectively have the highest likelihood and lowest distance to its assigned cluster (Figure S4). They also have low within-cluster variability as measured by the trace of sample covariance (Figure 3). On the other hand, days characterized with low likelihood scores and high distance scores tend to be more dispersed and do not conform well to a specific routine. For example, cluster 5 and cluster 7 in the GMM model and cluster 9 in the PAM model have such properties. Overall, the GMM clustering and PAM clustering tend to produce clusters with different behavioral representations in the assigned clusters and this is reflected in the clustering-based features such as likelihood scores and cluster distance that are assigned to characterize each day.

In terms of relapse prediction, clustering-based features can capture long-term behavioral trends across the subjects. This representation can complement existing approaches to behavioral representation for psychotic relapse prediction in schizophrenia, e.g., based on the usage of daily behavioral rhythm change features as proposed in [19]. We compared the clustering-based features before and near the relapse periods and saw significant differences in some of the features. This was also seen qualitatively in a time-series plot of clustering-based features indicating that an upcoming relapse for a patient is associated with changes seen in clustering-based features (Figure 5). Clustering-based features were helpful in relapse prediction models (Table 3). When clustering-based features were used together with daily behavioral rhythm change features, a significant gain in relapse prediction performance was obtained (F2 score improved from 0.18 to 0.23). These F2 scores, and the associated improvements are significant, considering that a random classification baseline gives an F2 score of 0.042 on average. A Wilcoxon signed-rank test on performances in multiple classifier initializations for classification with and without clustering features yielded a significant classification score for classification when clustering features were included ($p = 0.002$). Clustering-based features were among the top features when significance of features for the relapse prediction task was evaluated (Table 4). Features such as mean cluster labels and number of transitions of labels were among the top (most frequently selected) features. Thus, both the information about which behavioral clusters the observations from the current period of monitoring belong to (likely representing behavioral clusters that are not normal behaviors) and how often transitions between different behavioral clusters happen (representing the patient showing frequent behavioral variations) are likely predictive of an oncoming relapse.Clustering-based features alone also proved to be valuable for relapse prediction. GMM-based and PAM-based clustering features only used in the relapse prediction pipeline led to an F2 score of 0.16 and 0.16 for relapse prediction respectively (Table 3). Therefore, clustering-based features are found to be a useful approach to obtain behavioral representations and can be employed in clinical applications such as relapse prediction.

## Comparison to previous work, limitations, and future research

To our best knowledge, this is the first work that used clustering analysis to group behavioral patterns of individuals with schizophrenia. Compared to previous works that used the hourly data to train the relapse prediction models, our work based on clustering features to represent different behavioral patterns has better model interpretability. Clustering analysis allows clinicians to understand different types of patient routines, as well as their frequencies. In terms of schizophrenia, cluster transitions observed before relapses could suggest which types of behavior are potential relapse related behavioral signatures. Intervention strategies to prevent relapses can then be made accordingly.

Researchers have studied how missing data is related to relapses and anomalies in mental health conditions. In the dataset that we have for evaluation in this work, some passive sensing daily templates have consecutive hours with missing data from almost all signal modalities. While in an anomaly detection study, Adler et al. used mean imputation [21], here we filled missing values with zeros. Filling missing data with mean values might ignore the potential relationship between missing data and anomalies. In reality, it is highly possible that out-patients may turn off their phone when they experience relapse symptoms. We observed that indeed there are more days from an inactive sensor reading cluster closer to relapses. The increased prevalence of inactive days also caused likelihood scores to increase and distance scores to decrease before relapses. Initially, we hypothesized that adhering to any routine or any cluster center might reduce the risk of relapse, but it turned out that some routines, such as missing sensing data, is actually associated with a higher risk of relapse.

Although the clustering features successfully improved relapse prediction results, the only observable relapse signature is an increase in likelihood score or a decrease in distance score, and a transition to an inactive cluster. For the relapse events that were not indicated by sensor inactiveness, we did not find any non-trivial changes in any specific feature prior to the relapse. Similarly, the relapse prediction performance with the best F2 score of 0.23 is relatively low. However, investigations of mobile sensing based relapse prediction in mental health disorders are relatively new and further improvements in this field could be expected as more dataset become available and improvements in machine learning models to the specific task of relapse prediction. In [43], relapse prediction in bipolar disorder was developed using clinical assessment features during patient visits. A high F score (F1) of upto 0.80 was reported. The relapse rate was quite high (relapse in >60% of the included patient) in the dataset used by the authors and the relapse prediction was done on a patient level (instead of a weekly prediction model in free-living conditions as in our case) which might have led to higher performance.

In this work, we obtained patient-independent clusters i.e. generalized behavioral clusters by pooling data from all the patients. We generalized that there are certain types of routines across all outpatients with schizophrenia. Future studies can focus on establishing personalized cluster models. As suggested in [25], every patient's relapse signatures and the extent to which they adhere to their daily routine are different. The same study found that individual-level models could achieve better performance in predicting symptom severity. Our model also found that participants have different routines as their frequency of staying in different clusters largely varies. Moreover, although most patients had higher likelihood scores and lower distance scores closer to relapses, some other patients demonstrated the opposite trend. Generalized behavioral models might not fully represent and discount the effect of different confounding variables such as job type, gender, current health, etc. that could impact behavioral trends. Though we used model personalization in relapse prediction, only the factor of age as a covariate of behavioral trends was considered. Personalized cluster models that account for different aspects of interpersonal differences would further help mitigate possible biases in behavioral representations due to confounding variables. Personalized relapse prediction models will also be required to test the effectiveness of the individual-level clusters. However, sufficient data for each new patient is needed to find cluster models specific for the patient and thus clinical deployment for new patients will be delayed. Cluster adaptation from generalized cluster models to personalized cluster models as more patient-specific data becomes available needs to be investigated in future work.

## Conclusion

In this work, we proposed a methodology to compute clustering models on 24-hour daily behavior of schizophrenia outpatients and showed that information extracted from the cluster model improved relapse prediction. New features were generated from the cluster models by measuring every observation's deviation from the cluster centers representing typical behavioral patterns. Two different clustering models were investigated. The GMM model allows for cluster overlap and has a more extreme cluster dispersion. The PAM model with DTW distance creates partitional clusters that are more generalized towards new data but fails to identify dense clusters. The clustering-based features in addition to the baseline features helped to improve relapse prediction model performance. In future work, we will further investigate personalized clusters and relapse prediction models.

# Abbreviations

GMM: Gaussian mixture model

PAM: Partition Around Medoids
DTW: Dynamic time warping
BRF: Balanced Random Forest
PCA: Principal component analysis
EM: Expectation-maximization
AIC: Akaike information criterion
BIC: Bayesian information criterion
NRx: x-day near relapse period

## GMM Model Illustration

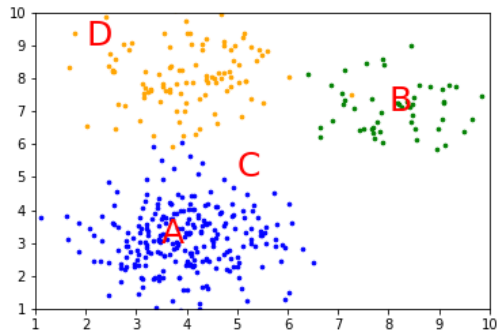

Supplementary Figure 1: 2-D illustration of Gaussian mixture and likelihood scores (cluster likelihood and weighted average likelihood) computed for different example points. Point A and Point B have similar cluster likelihood scores as they lie near their respective cluster centers but Point A has higher weighted average likelihood as it belongs to a larger cluster. Point C and Point D have similarly lower cluster likelihood scores (lying farther from their closest cluster centers) but Point D has the lowest average likelihood score as it is farthest away from all the cluster centers. An anomalous behavior (a day with atypical behavior) will likely resemble Point D.

## GMM Model Selection

Two parameters needed to be selected for the GMM model: the number of clusters $k$ (ranging from 5 to 20) and the covariance matrix type (spherical, tied, diagonal, full). First we computed Akaike information criterion (AIC) and Bayesian information criterion (BIC) scores of all the candidate models (obtained using all the combinations of the possible covariance matrix types and the number of clusters). A main issue exists when training a GMM model, as also observed in our evaluation. Large cluster overlaps between gaussian components with low and high variance could be obtained, leading to non-informative clusters as it fails to identify a specific routine. To address this problem, for each model selection, we computed the average pairwise Bhattacharyya distances [1] between all fitted Gaussian distributions to evaluate the difference between the generated clusters. Larger value indicates larger separation and less overlap between clusters. The k-means++ [2] approach was utilized to determine the model initialization condition. This approach randomly selected the first centroid, computed clusters, and assigned the farthest point in the cluster model to be the next centroid, until $k$ centroids were found, which were used as the initial centroids of the GMM model. To evaluate model stability, each model selection was trained five times with different randomizations. The mean and standard error of each metric was illustrated in Supplementary Figure 2. The optimal GMM model selected has high stability, low overlap, low AIC, and BIC.

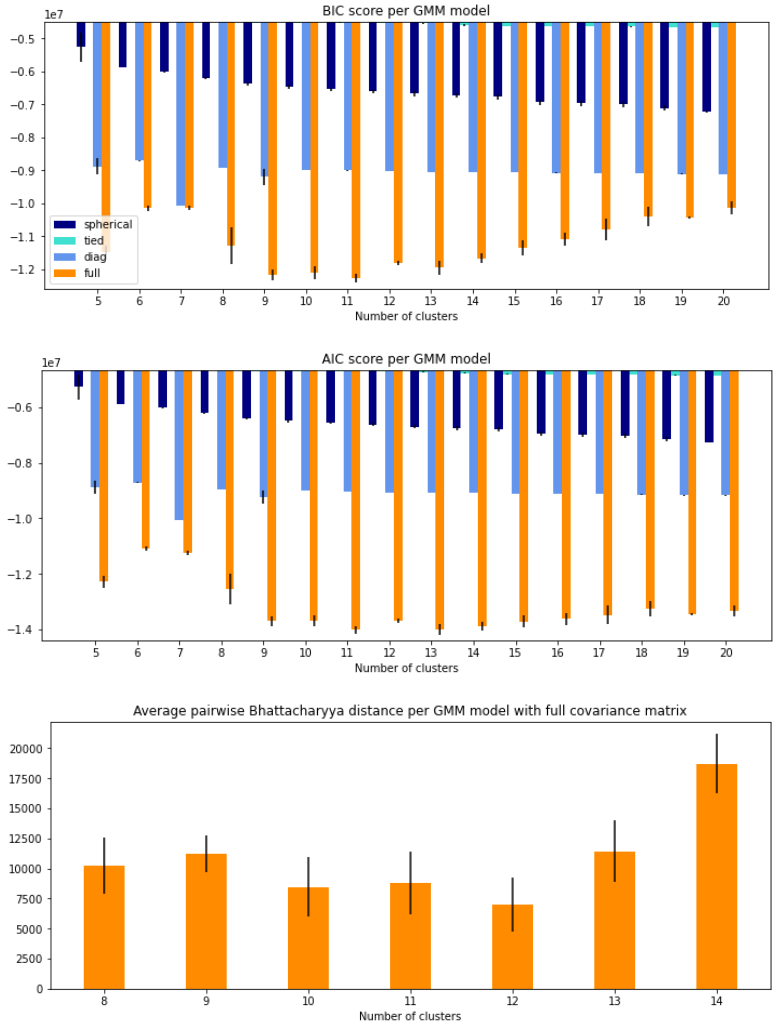

Supplementary Figure 2: BIC scores, AIC scores, and average pairwise Bhattacharyya distances for each GMM model selection.

## PAM Model Selection

We considered the number of clusters *k* ranging from 5 to 20, in the same range as those considered within the GMM model. For each possible number of clusters, we trained the model five times with different initializations, and computed the sum of the squared DTW distance of every data point to its cluster medoid. The average sum of squared distance and the standard error were plotted in Supplementary Figure 3. The final model was selected using the elbow method, which means the cluster number is optimal when adding one more cluster does not contribute too much to reducing the sum of squared distance.

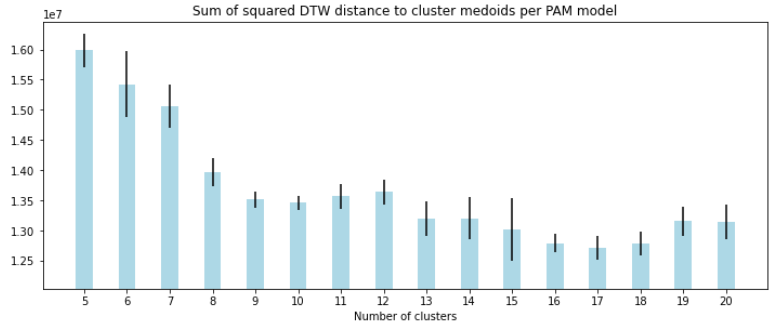

Supplementary Figure 3: sum of squared DTW distance for each PAM model selection

## Relapse Prediction: Hyperparameters for the classifier

Our relapse prediction pipeline using the Balanced Random Forest (BRF) as classifier has a few hyper-parameters namely: the number of bins to quantize features, size of personalization subset, and number of features to be selected for model training. We set the value of these hyper-parameters based on the cross-validation results obtained with different sets of parameter values within the training set. A nested cross-validation with random k-fold (k=10) within the training set was used to identify the hyper-parameter to be used for a given patient in the test set. We used a random k-fold cross-validation in the inner loop as leave-one-patient-out cross-validation would be computationally costly (number of folds would be 61 in the inner loop also). A random k-fold partition in the inner loop could still be helpful to identify good hyperparameters when a higher number of folds are used since most of the subjects would appear in the test set in one of the folds and the discovered hyperparameters are thus generalizable. The parameter values considered for identifying the best parameters were: [2, 3, 4, 5, 10, 15] for the number of bins, [50, 75, 100, 125, 150, 200, 300] for the size of personalization subset, and [3, 5, 10, 15] for the number of features to be selected.

## Clustering and Relapse Prediction Results

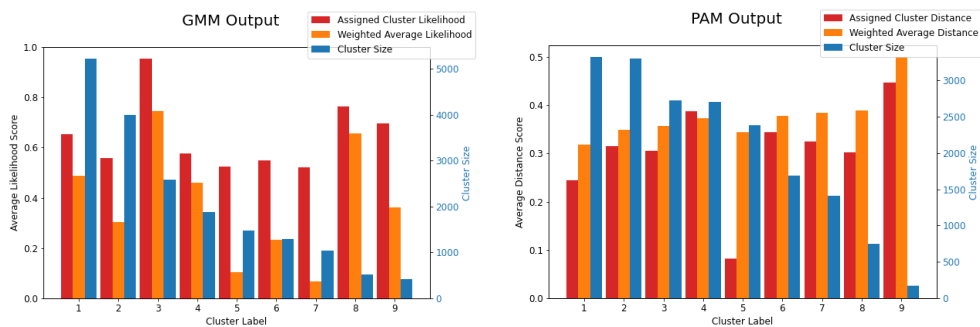

Supplementary Figure 4.A (left): average assigned cluster likelihood score, average weighted average likelihood score, and cluster size of each GMM model cluster. Figure 4.B (right): average assigned cluster distance score, average weighted average distance score, and cluster size of each PAM model cluster.

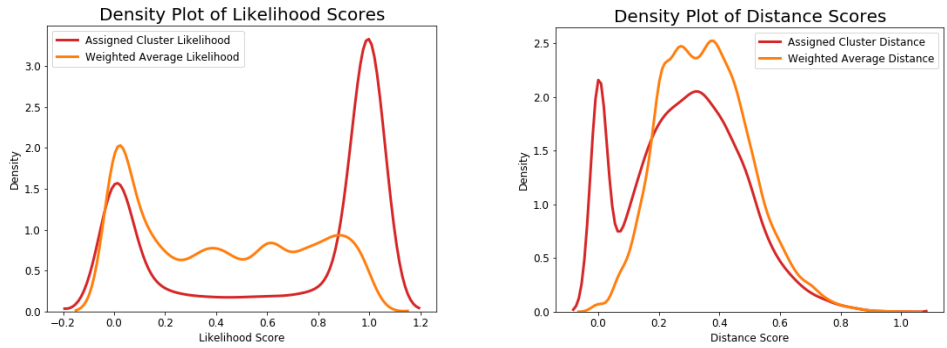

Supplementary Figure 5.A (left): kernel density plot of assigned cluster likelihood score and weighted average likelihood score from the GMM model for all sample data. Figure 5.B (right): kernel density plot of assigned cluster distance score and weighted average distance score from the PAM model for all sample data.

|  | NR7 vs. pre-NR7 | NR14 vs. pre-NR14 | NR20 vs. pre-NR20 | NR30 vs. pre-NR30 |
|---|---|---|---|---|
| Assigned cluster likelihood (cliff's delta) | 0.1346 | 0.1681 | 0.1963 | 0.1129 |
| Weighted average likelihood (cliff's delta) | 0.2023 | 0.2416 | 0.2552 | 0.2473 |
| Assigned cluster distance (cliff's delta) | -0.1598 | -0.2150 | -0.2504 | -0.2386 |
| Weighted average distance (cliff's delta) | -0.1787 | -0.1862 | -0.2098 | -0.2171 |

Supplementary Table 1: differences in clustering features between the NRx (x days near relapse) and the pre-NRx (all days before relapses not in NRx) periods.

|  | F2 score | |
|---|---|---|
| Method | With personalization (age-based) | Without personalization |
| All features | 0.23 | 0.14 |
| Baseline features | 0.18 | 0.14 |
| Clustering features | 0.14 | 0.11 |
| GMM features | 0.16 | 0.13 |
| PAM features | 0.16 | 0.16 |
| GMM + Baseline features | 0.19 | 0.13 |
| PAM + Baseline features | 0.16 | 0.14 |

Supplementary Table 2: Effect of age-based personalization on relapse prediction performance. A higher F2 score in relapse prediction is obtained when age-based personalization is used compared to when no personalization is used.